\documentclass{article}
\usepackage{ijcai23}

\usepackage{times}
\usepackage{soul}
\usepackage{url}
\usepackage[hidelinks]{hyperref}
\usepackage[utf8]{inputenc}
\usepackage[small]{caption}
\usepackage{graphicx}
\usepackage{amsmath}
\usepackage{amsthm}
\usepackage{booktabs}
\usepackage[ruled,vlined]{algorithm2e}
\usepackage{threeparttable}
\usepackage[noend]{algpseudocode}
\usepackage{amssymb}
\usepackage[switch]{lineno}

\title{Fast-StrucTexT: An Efficient Hourglass Transformer with Modality-guided Dynamic Token Merge for Document Understanding}

\author{
    Mingliang Zhai$^{1,2}$ \and
    Yulin Li$^{2}$ \and
    Xiameng Qin$^{2}$ \and
    Chen Yi$^{2}$ \and
    Qunyi Xie$^{2}$ \and
    Chengquan Zhang$^{2}$ \and
    Kun Yao$^{2}$ \and
    Yuwei Wu$^{1,3}$ \and
    Yunde Jia$^{1,3}$ \and
    \affiliations
    $^1$Beijing Key Laboratory of Intelligent Information Technology, School of Computer Science \& Technology, Beijing Institute of Technology, China\\
    $^2$Department of Computer Vision Technology (VIS), Baidu Inc\\
    $^3$Guangdong Laboratory of Machine Perception and Intelligent Computing, Shenzhen MSU-BIT University, China
    \emails
    \{zhaimingliang, wuyuwei, jiayunde\}@bit.edu.cn, \\
    \{liyulin03, qinxiameng, chenyi17, xiequnyi, zhangchengquan, yaokun01\}@baidu.com
}
\begin{document}
\maketitle

\begin{abstract}

Transformers achieve promising performance in document understanding because of their high effectiveness and still suffer from quadratic computational complexity dependency on the sequence length. General efficient transformers are challenging to be directly adapted to model document. They are unable to handle the layout representation in documents, \emph{e.g.} word, line and paragraph, on different granularity levels and seem hard to achieve a good trade-off between efficiency and performance. To tackle the concerns, we propose \textbf{Fast-StrucTexT}, an efficient multi-modal framework based on the StrucTexT algorithm with an hourglass transformer architecture, for visual document understanding. Specifically, we design a modality-guided dynamic token merging block to make the model learn multi-granularity representation and prunes redundant tokens. Additionally, we present a multi-modal interaction module called Symmetry Cross Attention (SCA) to consider multi-modal fusion and efficiently guide the token mergence. The SCA allows one modality input as query to calculate cross attention with another modality in a dual phase. Extensive experiments on FUNSD, SROIE, and CORD datasets demonstrate that our model achieves the state-of-the-art performance and almost 1.9$\times$ faster inference time than the state-of-the-art methods.

\end{abstract}

\section{Introduction}
\label{sec:intro}

Visual document understanding technology aims to analyze visually rich documents (VRDs), such as document images or digital-born documents, enables to extract key-value pairs, tables, and other key structured data.
At present, multi-modal pre-training transformer models~\cite{ChenYuLee2022FormNetSE,gu2022xylayoutlm,huang2022layoutlmv3,hong2022bros} have shown impressive performance in visual document understanding. 
The inside self-attention mechanism is crucial in modeling the long-range dependencies to capture contextual information. 
However, its quadratic computational complexity is the limitation of the transformer involved in visual document understanding with long sequences directly.

\begin{figure}
    \centering
    \includegraphics[width=1.0\linewidth]{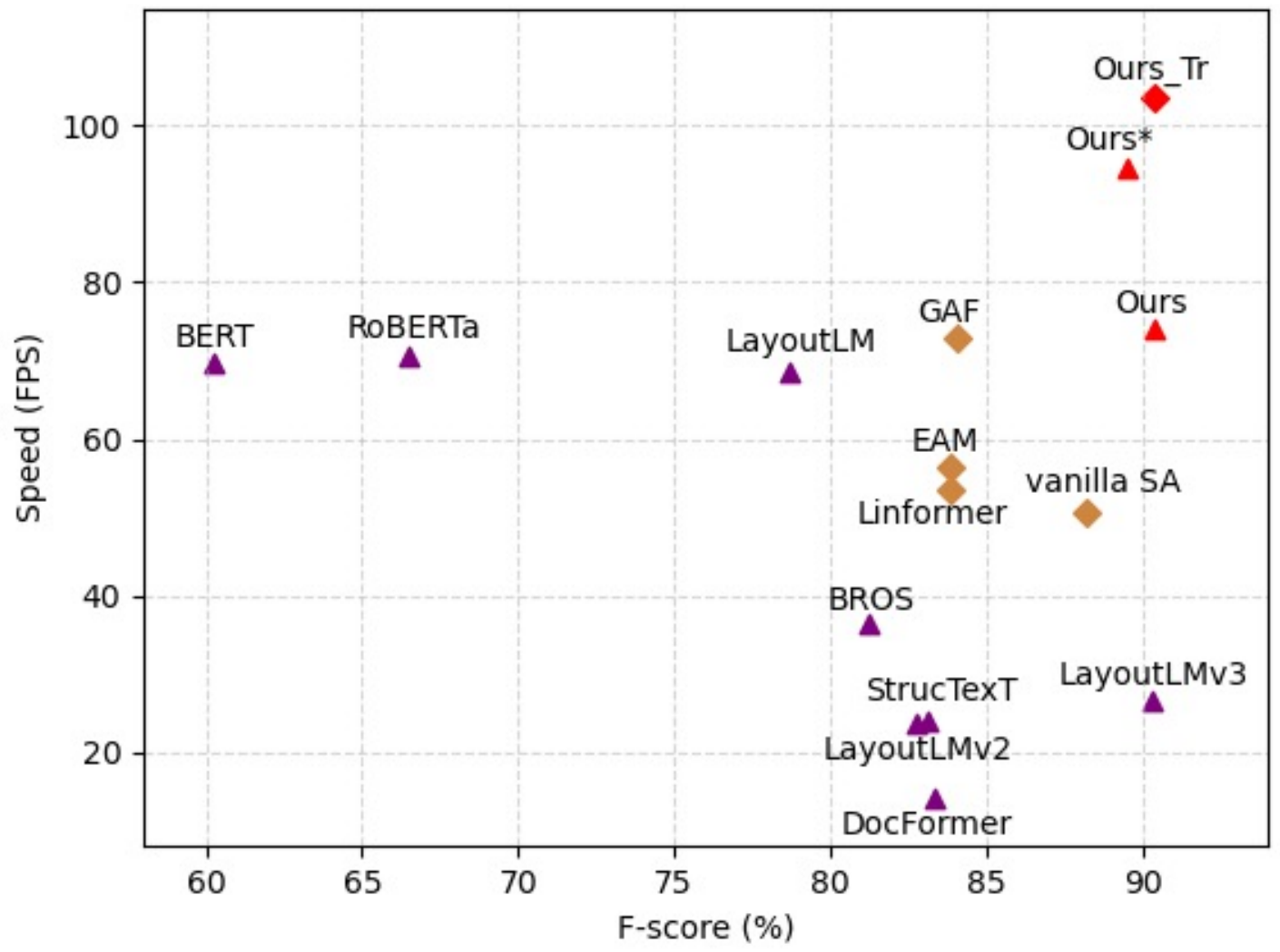}
    \caption{F-score (\%) vs Inference Speed (FPS) on FUNSD test set. \textit{Triangles} indicate the methods focus on document understanding, and the red triangle is our method. \textit{Diamonds} mean that we utilize the efficient transformers in our framework, and the FPS is computed without image embeddings. The red diamond shows our method.}
    \label{fig:speed}
\end{figure}

\begin{figure*}
    \centering
    \includegraphics[width=0.99\linewidth]{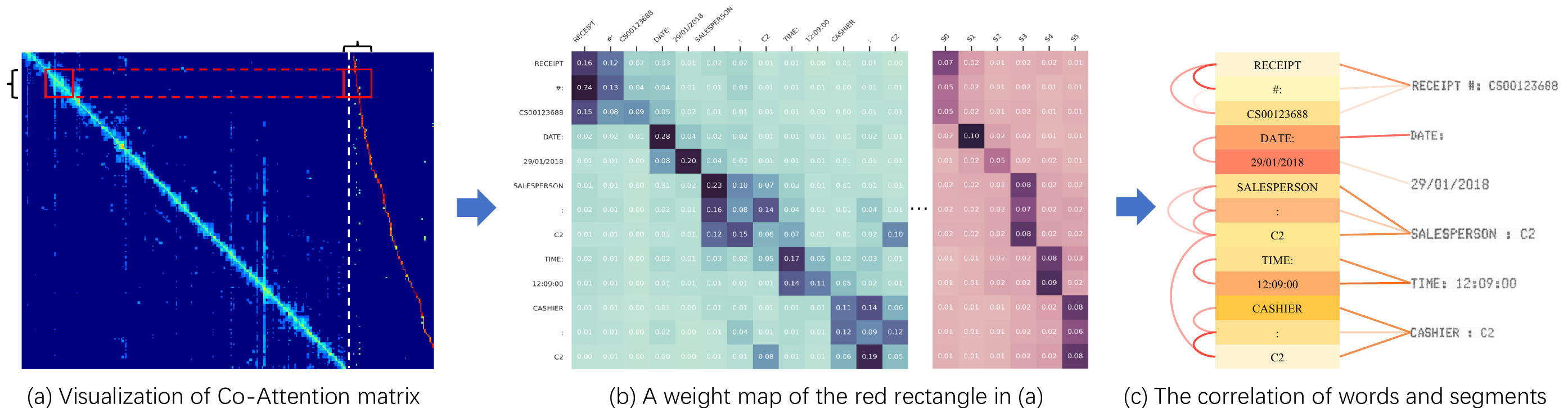}
    \caption{We accumulate the attention map of all layers to get (a), and then enlarge the part in the red box to get (b). According to the word and segment corresponding to this area, we get (c). The line represents the correlation between words and segments.}
    \label{fig:weights}
\end{figure*}

To improve the computational efficiency, there are three typical solutions in general. 
The first solution~\cite{SehoonKim2021LearnedTP,yin2022vit} simply reduces input sequence length by efficient sampling or new tokenization process.
The second solution attempts to redesign transformer architectures, such as reformulating the self-attention mechanism~\cite{SinongWang2020LinformerSW,JianWang2022RTFormerED,IzBeltagy2020LongformerTL} or producing a lightweight model with a smaller size~\cite{XiaoqiJiao2019TinyBERTDB,KanWu2022TinyViTFP,JinnianZhang2022MiniViTCV}. 
The third solution combines tokens by MLP~\cite{MichaelSRyoo2021TokenLearnerWC}, grouping~\cite{JiaruiXu2022GroupViTSS}, or clustering~\cite{DmitriiMarin2021TokenPI} to prune unnecessary tokens related to several works~\cite{JuntingPan2022EdgeViTsCL,IzBeltagy2020LongformerTL} have proved the redundancies in attention maps.

Although those methods significantly reduce the computational complexity of the transformer model, they do not take into account the multiple granularity expressions in the visual documents, including words, lines, paragraphs, and so on. 
Furthermore, some token reduction methods learn multi-granularity information to a certain extent, but without considering the correlation between modality and granularity.

In Figure~\ref{fig:weights}, we visualize the attention maps and token interactions of a standard transformer in the inference phase. In particular, Figure~\ref{fig:weights}(a) displays the aggregated attention map of all Transformer layers. The left half shows a self-attention map inside words and the right half shows the cross-attention map between words and segments. We re-sample two regions with highlighted red boxes and zoomed in Figure~\ref{fig:weights}(b). Moreover, for closer observation, Figure~\ref{fig:weights}(c) gives the correlation visualization (curves and lines) of words and segments based on attention scores related to Figure~\ref{fig:weights}(b). Almost the words and segments belonging to one semantic entity have significant correlations to their counterparts, indicating a high redundancy between irrelevant tokens in attention scores, which illustrates two key viewpoints:
\begin{enumerate}
\item Strong correlations are existed across granularities.
\item There is rich redundancy in attention computation.
\end{enumerate}

\begin{figure*}
    \centering
    \includegraphics[width=1.0\linewidth]{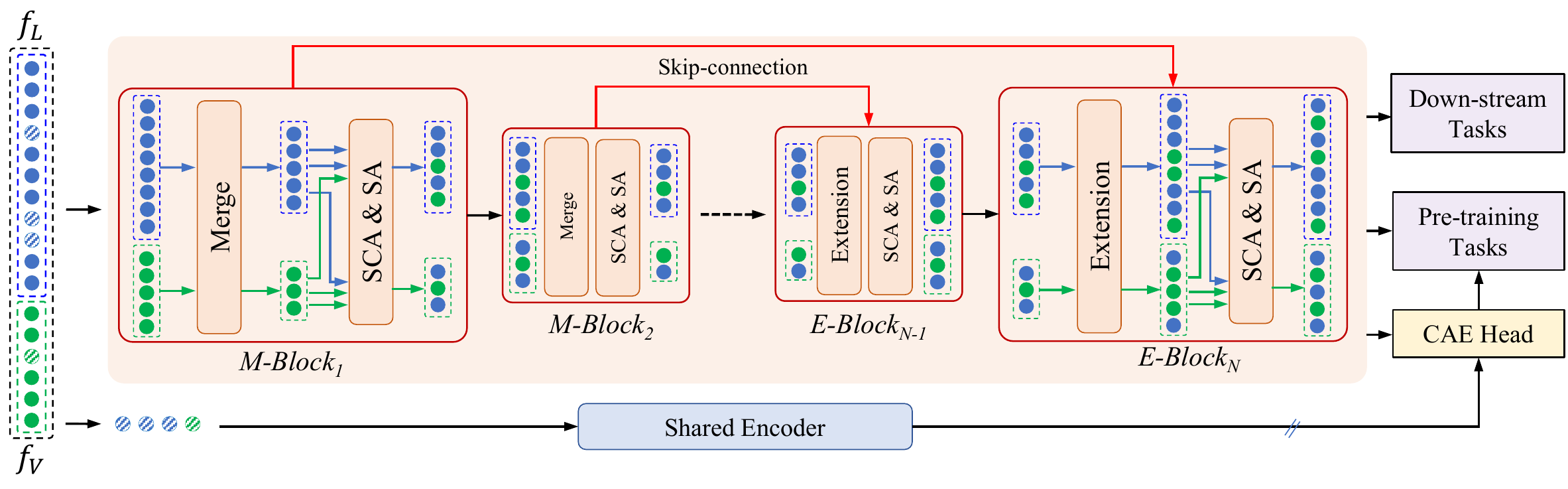}
    \caption{The overall Fast-StrucText architecture. Encoder is an hourglass architecture with the input of language and visual features. The merging operation can effectively reduce redundancy computations, and multi-modal interaction information is obtained through SCA \& SA. The skip-connection exits from before merging to after extension.}
    \label{fig:architecture}
\end{figure*}

In this paper, we propose an efficient multi-modal transformer called \textbf{Fast-StrucTexT} based on StrucTexT~\cite{li2021structext}, which is not only devoted to improving the model efficiency but also enhancing the expressiveness.
% hourglass
We design an hourglass transformer that consists of merging and extension blocks, which receives the full-length sequence as input and compresses the redundant tokens progressively. A weighted correlation between two granularity features is produced by each merging block to dynamically guide the merging process. Since massive redundant information is eliminated, our model yields great efficiency gain and higher-level semantics.
% upsample
To address the problem that some downstream tasks, such as named entity recognition (NER), require a sufficient number of tokens, we decode the shortened hidden states to the full-length sequence. This hourglass architecture can take full of each token representation and use fewer tokens to complete semantics. Since the number of tokens has been reduced in the middle layers of transformer, the efficiency of the model has been greatly improved.

% SCA
It is vital that multi-modal interaction for guiding merge and semantic representation. We develop a Symmetry Cross-Attention (SCA) module, a dual cross-attention mechanism with multi-modal interaction. One model feature is used as the query and the other modal feature is used as the key and value to calculate cross-attention in the visual and textual modalities respectively. Significantly, SCA and Self-Attention (SA) are used alternately in our architecture. The SA provides global modal interaction and the SCA conducts multi-modal feature fusion and provide modal semantic guidance for token merging.

We pre-train Fast-StrucTexT with four task-agnostic self-supervised tasks for learning a good representation, and then fine-tune the pre-trained model in three benchmark datasets. Experiment results demonstrate that our model achieves state-of-the-art performance and FPS.

The main contributions are summarized below:
\begin{itemize}
\item We propose Fast-StrucTexT with an efficient hourglass transformer by performing modal-guided dynamic token merging to reduce the number of tokens.
\item We develop a dual multi-modal interaction mechanism named Symmetry Cross-Attention, which can enhance the multi-modal feature fusion from visual documents.
\item Extensive experiments on four benchmarks show our model achieve state-of-the-art speed and performance.
\end{itemize}

% ---------------------------- Related Work ----------------------------
\section{Related Work}

\subsection{Multi-Modal Pre-training Model}
% Document understanding has made rapid progress due to multi-modal pre-training representation learning. 
As the first heuristic work, NLP-based approaches~\cite{JacobDevlin2018BERTPO,YinhanLiu2019RoBERTaAR} adopt the language model to extract the semantic structure. Various works~\cite{hong2022bros,ChenYuLee2022FormNetSE,li2021structurallm,xu2020layoutlm} then jointly leverage layout information by spatial coordinates encoding, leading to better performance and extra computations simultaneously.
After that, some researchers realize the effectiveness of deep fusion among textual, visual, and layout information from document images. A quantity of works~\cite{JiuxiangGu2021UniDocUP,li2021selfdoc,li2021structext,xu2020layoutlmv2,gu2022xylayoutlm,appalaraju2021docformer} rely on text spotting to extract semantic region features with a visual extractor.
% A range of methods, such as LayoutLMv2~\cite{xu2020layoutlmv2}, XYLayoutLM~\cite{gu2022xylayoutlm}, and DocFormer~\cite{appalaraju2021docformer}, explore grid-based image features with patch embeddings. 
% Benefiting from the existing OCR engines, several  
% All those image representation learning approaches would account for heavy computation bottlenecks.
However, the transformer-based architectures are inefficient for long sequence modeling because of computationally expensive self-attention operations.
% transformer architecture still takes the main responsibility for parameters and flops when it is faced with a long sequence input.
To this end, we propose an efficient multi-modal model, Fast-StrucTexT, for visual document understanding.
% Despite patch learning with a light-weight linear layer~\cite{huang2022layoutlmv3}, the transformer module still takes the main responsibility for parameters and flops when it is faced with a long sequence. 

\subsection{Efficient Transformers}
A well-known issue with self-attention is its quadratic time and memory complexity, which can impede the scalability of the transformer model in long sequence settings. Recently, there has been an overwhelming influx of model variants proposed to address this problem.

\subsubsection{Fixed Patterns}
The earliest modification to self-attention simply specifies the attention matrix by limiting the field of view, such as local windows and block patterns of fixed strides. Sparse Transformer~\cite{RewonChild2019GeneratingLS} converts the dense attention matrix to a sparse version by only computing attention on a sparse number of $q_i, k_j$ pairs. GroupingViT~\cite{JiaruiXu2022GroupViTSS} divides tokens into multiple groups, and then group-wised aggregates these tokens. Chunking input sequences into blocks that reduces the complexity from $N^2$ to $B^2$ (block size) with $B \ll N$, significantly decreasing the cost.

\subsubsection{Down-sampling}
Down-sampling methods that narrow the resolution of a sequence can effectively reduce the computation costs by a commensurate factor. Zihang Dai \emph{et al.}~\cite{ZihangDai2020FunnelTransformerFO} have highlighted the much-overlooked redundancy in maintaining a full-length token-level representation. To solve this problem, they compress the sequence of hidden states to a shorter length, thereby reducing the computation cost. TokenLearner~\cite{MichaelSRyoo2021TokenLearnerWC} uses MLP to project the tokens to low-rank space. The recent Nystr\"omformer~\cite{YunyangXiong2021NystrmformerAN} is a down-sampling method in which the "landmarks" are simply strided-based pooling in a similar spirit to Set Transformer~\cite{lee2019set}, Funnel Transformer~\cite{ZihangDai2020FunnelTransformerFO}, or Perceiver~\cite{jaegle2021perceiver}. Inspired by those works, we design the hourglass transformer composed of modality-guided dynamic token merging and extension.

% ---------------------------- Method ----------------------------
\section{Method}
In this section, we provide the framework of the proposed Fast-StrucTexT. First, we introduce the model architecture and describe the approach for generating multi-modal input features. Next, we present the details of the hourglass transformer with a hierarchical architecture. Finally, we explain the pre-training objectives and fine-tuning tasks. The overall Fast-StrucTexT architecture is depicted in Figure~\ref{fig:architecture}.

\subsection{Model Architecture}
\label{section:3.1}
Given a visual document and its text content extracted from OCR toolkits, the feature extractor first extracts both textual and visual features from the region proposals of text segments. 
These features are then fed into a transformer-based encoder to learn multi-modal contextualized representations via alternating self-attention and cross-attention mechanisms. 
By leveraging redundancies across input tokens, the proposed encoder consisting of several Merging-Blocks and Extension-Blocks is designed as a hierarchical structure for shortening the sequence length progressively. In particular, the Merging-Block dynamically merges nearby tokens and the Extension-Block recovers the shortened sequence to the original scale according to the merging information to support token-level downstream tasks. 
Besides, we further adopt the alignment constraint strategy to improve the model ability by introducing a CAE~\cite{XiaokangChen2022ContextAF} head in the pre-training. The generated contextual representations can be used for fine-tuning downstream tasks of visual document understanding.

\subsection{Feature Extraction}
\label{section:3.2}
We employ an off-the-shelf OCR toolkit to a document image $I \in \mathbb{R}^{W \times H}$ to obtain text segments with a list of sentences $S$ and corresponding 2D coordinates $B$. We then extract both segment-level visual features and word-level textual features through a ConvNet and a word embedding layer. For visual features $f_V$, the pooled RoI features of each text segment extracted by RoIAlign are projected to a vector. For textual features $f_L$, we utilize the WordPiece to tokenize text as sub-words and convert them to the ids. 
% In addition, following the LayoutLMv3, the position embeddings include 1D position and normalized 2D layout position embeddings are appended to the respective $f_V$ and $f_L$. 
% The 1D position refers to the index from 1 to maximum token length and the 2D layout position encodes the coordinates of text segments.

% We start by extracting textual and visual features from region proposals of text segments, which serve as input to a feature extractor. 
% The resulting features are then passed to a transformer-based encoder that learns multi-modal contextualized representations using alternating self-attention and cross-attention mechanisms. 
% We designed the encoder as a hierarchical model that shortens the sequence length by merging nearby tokens into higher semantic components through M-Blocks. 
% The E-Blocks then recover the fused sequence to the original scale to support common pre-training tasks such as masked language modeling (MLM). 
% In addition, we introduce an alignment constraint strategy using a CAE Head~\cite{XiaokangChen2022ContextAF} to improve the capacity of model pre-training. 
% The resulting contextual representations can be used for fine-tuning downstream tasks related to document understanding.

We add a special start tag \texttt{[CLS]} with coordinates $B_0=[0, 0, W, H]$ at the beginning of the input to describe the whole image. Besides, several \texttt{[PAD]} tags with zero bounding boxes append to the end of $f_V$, $f_L$ to a fixed length. We ensure the length of $f_L$ is an integral multiple of $f_V$.

\subsection{Hourglass Encoder}
\label{section:3.3}
% Inspired by the hierarchical architectures in CNN~\cite{XuT21} and Transformer~\cite{NawrotTTKWSM22}, we
We propose an hourglass transformer as the encoder module and progressively reduce the redundancy tokens. Instead of shortening the sequence length by a fixed pattern of merging tokens, the model performs a dynamic pooling operation under the guidance of semantic units (text sentence/segment), which leads to a shorter hidden states without harming model capability. The encoder consists of several Merging- and Extension-blocks. The Merging-block (M-Block) merges the tokens of hidden states and the Extension-block (E-Block) conducts up-sampling to make up for the original length.

\begin{figure}
    \centering
    \includegraphics[width=1.0\linewidth]{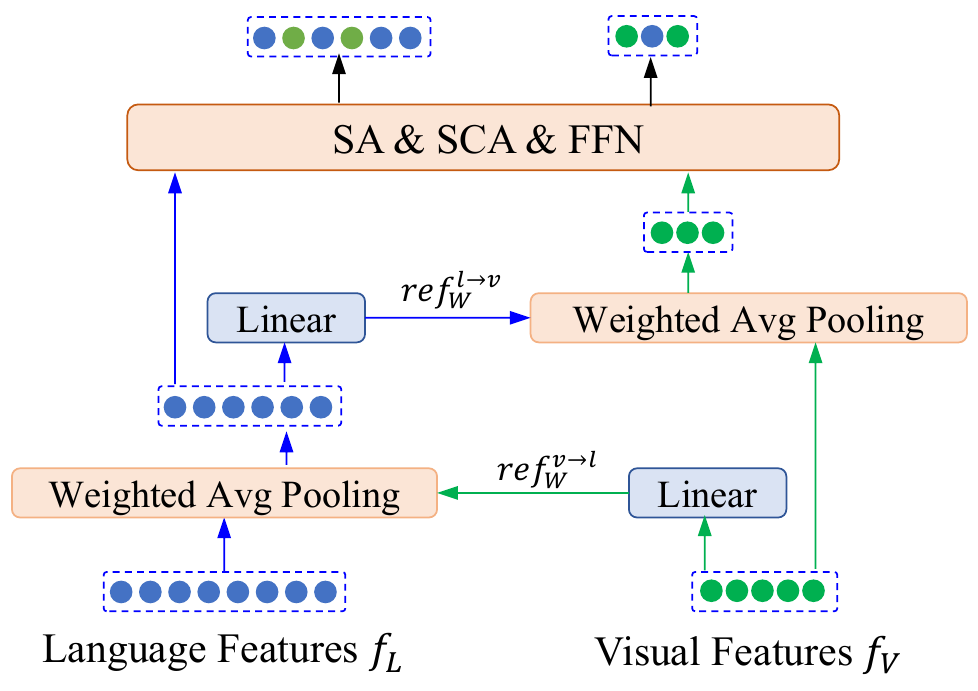}
    \caption{Proposed modality-guided dynamic token merge block.
    }
    \label{fig:merge}
\end{figure}

\textbf{Merging}.
The M-Block suggests merging nearby $k$ tokens with weighted 1D average pooling, where $k$ is the shortening factor. In view of the multi-model hidden states, a referred weighting is predicted from another modality by a linear layer. The process is denoted in Figure~\ref{fig:merge}.

\begin{equation}
f^i_n = \text{AVG\_POOL}(f^{i-1}_n \cdot \text{Linear}^{m \rightarrow n}(f^{i-1}_m)),
\end{equation}
where $i$ is the stage index of M-Block, $m$ and $n$ denotes textual and visual modality $(m, n \in \{V,L\})$. Notably, $\text{Linear}^{* \rightarrow *}$ share their parameters for all M-Blocks.

\textbf{Extension}.
The E-Block is required to transform shortened sequence of hidden states back to the entire token-level state. In detail, we simply apply repeat up-sampling method~\cite{DaiLY020} to duplicate the vector of each merged token for $k$ times. This method is computationally efficient. For maintaining the distinct information of tokens, we fed the hidden states from the corresponding M-Block into the E-Block through a skip-connection.

\textbf{Symmetry Cross-Attention}.
Cross-Attention has shown the effectiveness of multi-modal learning on the vision-and-language tasks. To enhance the interactions among textual and visual modalities, we introduce Symmetry Cross-Attention (SCA) module to model the cross-modality relationships, which consists of two dual cross-attentions to handle the text and visual features in this work. We also leverage SCA to provide cross-modality guidance for token merging in the M-Blocks. The SCA is defined as follows:
\begin{equation}
\begin{aligned}
& \text{CA}(f_n|f_m) = W_o\sigma(\frac{F_q(f_n)F^T_k(f_m)}{\sqrt{d}})F_v(f_m) \\
& \text{SCA}(f_n, f_m) = \{\text{CA}(f_n|f_m), \text{CA}(f_m|f_n)\}
\end{aligned}
\end{equation}
where $F_q, F_k, F_v$ are the linear project for query, key and value. $\sigma$ is the \textit{Softmax} function, $d$ is the hidden size and $W_o$ is the output weight. The multi-head settings of attention are omitted for brevity. 
% Pseudo-codes of the SCA are Algorithm~\ref{alg:co-attention}.

Yet it's worth noting that our SCA incorporates semantic embedding as an additional input that gives the identical index for each segment and its corresponding words. SCA can provide multi-granularity interaction information for the subsequent token merging in addition to ensure the multi-modal interaction. Furthermore, a global context information be taken into account by Self-Attention (SA). Therefore, SCA and SA are adopted in turn to build transformer layers.

% \begin{algorithm}[t]
% \caption{Symmetry cross attention}
% \label{alg:co-attention}
% {\textbf{Input:}} The features of language tokens $f_L$ and features of visual tokens $f_V$;\\
% {\textbf{Output:}} The results of the multi-modal interaction;
% \Statex\qquad\STATE $F_V$, $F_L$ = Split($O_s$, [Len($F_V$), Len($F_L$)]);
% \Statex\qquad\STATE $Q_V$,$K_V$,$V_V$ = LinearProj($F_V$);
% \Statex\qquad\STATE $Q_L$,$K_L$,$V_L$ = LinearProj($F_L$);
% \Statex\qquad\STATE $O_V$ = Cross-attention($Q_V$, $K_L$, $V_L$);
% \Statex\qquad\STATE $O_L$ = Cross-attention($Q_L$, $K_V$, $V_V$);\\
% \STATE \Return Output = $W_o \cdot \text{Concat}(O_V, O_L)$;
% \end{algorithm}

\begin{table*}[t]
  % \begin{threeparttable}
  \begin{center}
  \begin{tabular}{@{\extracolsep{.25cm}}l @{}l @{}l @{}c @{}c @{}c @{}c @{}c @{}c}
    \toprule
    Model & Modality & Image & FPS$^\dagger$ & FLOPs & Param. & FUNSD & CORD & SROIE \\
          &          & Embedding  &               & (G)   &  (M)  & F1    & F1   & F1 \\
    \midrule
    $\text{BERT}_{BASE}$~\cite{JacobDevlin2018BERTPO} & T & None & 69.77 & 48.36 & \textbf{110} & 60.26 & 89.68 & 93.67 \\
    $\text{RoBERTa}_{BASE}$~\cite{YinhanLiu2019RoBERTaAR} & T & None & 70.47 & 48.36 & 125 & 66.48 & 93.54 & - \\
    $\text{LayoutLM}_{BASE}$~\cite{xu2020layoutlm} & T+L & None & 68.68 & 48.36 & 113 & 78.66 & 94.72 & 94.38 \\
    $\text{BROS}_{BASE}$~\cite{hong2022bros} & T+L & None & 36.29 & 54.00 & \textbf{110} & 81.21 & 95.36 & 95.48 \\
    $\text{FormNet}_{A2}$~\cite{ChenYuLee2022FormNetSE} & T+L & None & - & - & 217 & 84.69 & 97.10 & - \\
    $\text{StructuralLM}_{BASE}$~\cite{li2021structurallm} & T+L & None & - & - & 113 & 78.66 & -  & - \\
    % $\text{XDoc}$~\cite{chen2022xdoc} & T+L & None & - & - & 146 & 89.40 & - & - \\
    UniDoc~\cite{JiuxiangGu2021UniDocUP} & T+L+I & ResNet-50 & - & - & 272 & 87.96 & 96.64 & - \\
    $\text{StrucTexT}_{BASE}$~\cite{li2021structext} & T+L+I & ResNet-50 & 24.11 & 82.21 & 107$^\ddagger$ & 83.09 & - & 96.88 \\
    $\text{DocFormer}_{BASE}$~\cite{appalaraju2021docformer} & T+L+I & ResNet-50 & 14.32 & 93.13 & 183 & 83.34 & 96.33 & - \\
    SelfDoc~\cite{li2021selfdoc} & T+L+I & ResNeXt-101 & - & - & 137 & 83.36 & - & - \\
    $\text{LayoutLMv2}_{BASE}$~\cite{xu2020layoutlmv2} & T+L+I & ResNeXt-101 & 15.50 & 91.45 & 200 & 82.76 & 94.95 & 96.25 \\
    % $\text{XYLayoutLM}_{BASE}$~\cite{gu2022xylayoutlm} & T+L+I & ResNeXt-101 & - & - & - & 83.35 & - & - \\
    % $\text{ERNIE-Layout}_{BASE}$~\cite{ernie2.0} & T+L+I & ResNet-101 & - & - & 268 & 90.28 & 96.61 & \underline{97.19} \\
    $\text{LayoutLMv3}_{BASE}$~\cite{huang2022layoutlmv3} & T+L+I & Linear & 39.55 & 55.95 & 133 & \underline{90.29} & 96.56 & - \\
    \midrule
    $\textbf{Fast-StrucTexT}^*$ & T+L+I & Linear & \textbf{94.64} & \textbf{19.85} & \underline{111} & 89.50 & \underline{96.65} & 97.12 \\
    \textbf{Fast-StrucTexT} & T+L+I & ResNet-18 & \underline{74.12} & \underline{44.91} & 116 & \textbf{90.35} & \textbf{97.15} & \textbf{97.55} \\
    \bottomrule
    % \multicolumn{9}{l}{\footnotesize $^*$ The number of parameters except for visual backbone of image embedding.} \\
    % \multicolumn{9}{l}{\footnotesize $^{^\dagger}$ The number of parameters is calculated from the official released model.}
  \end{tabular}
  \end{center}
  \caption{Entity labeling performance and model efficiency comparison on FUNSD, CORD, and SROIE datasets. ``T/L/I'' denotes ``text/layout/image'' modality. FPS$^\dagger$ computation excludes the heads of downstream tasks. $^\ddagger$ indicates the parameter number of StrucTexT is calculated without visual backbone of image embedding.}
  \label{tab::labeling}
\end{table*}

\subsection{Pre-training Objectives}
\label{section:3.4}
We adopt four self-supervised tasks simultaneously during the pre-training stage, which are described as follows.
\textbf{Masked Visual-Language Modeling (MVLM)} is the same as LayoutLMv2~\cite{xu2020layoutlmv2} and StrucTexT~\cite{li2021structext}. Moreover, a CAE~\cite{XiaokangChen2022ContextAF} head is introduced to eliminate masked tokens in feature encoding and keep the consistency of document representation between pre-training and fine-tuning.

\textbf{Graph-base Token Relation (GTR)} constructs a ground truth matrix $G$ with 0$\sim$9 to express the spatial relationship between each pairwise text segments. We give $G_{ij}$ a layout knowledge, \emph{i.e.}, $0$ means the long distance (exceeding half the document size) between text segment $i$ and $j$, and 1$\sim$9 indicate eight buckets of positional relations (up, bottom, left, right, top-left, top-right, bottom-left, bottom-right). We apply a bilinear layer in the segment-level visual features to obtain the pairwise features and fed them into a linear classifier driven by a cross-entropy loss.

\textbf{Sentence Order Prediction (SOP)} uses two normal-order adjacent sentences as positive examples, and others as negative examples. SOP aims to learn fine-grained distinctions about discourse-level coherence properties. Hence, the encoder is able to focus on learning semantic knowledge, and avoids the influence of the decoder.

\textbf{Text-Image Alignment (TIA)} is a fine-grained cross-modal alignment task. In the TIA task, some image patches are randomly masked with zero pixel values and then the model is pre-trained to identity the masked image patches according to the corresponding textual information. It enable the combine information between visual and text. Masked text token is not participating when estimating TIA loss.

\subsection{Fine-tuning}
\label{section:3.5}
We fine-tune our model on visual information extraction tasks: entity labeling and entity linking.

\textbf{Entity Labeling}.
The entity labeling task aims to assign each identified entity a semantic label from a set of predefined categories. We perform an arithmetic average on the text tokens which belong to the same entity field and get the segment-level features of the text part. To yield richer semantic representations of entities containing the multi-modal information, we fuse the textual and visual features by the \textit{Hadamard} product operation to get the final context features. Finally, a fully-connected layer with \textit{Softmax} is built above the fused features to predict the category for each entity field. 
The Expression is shown as follow, 
\begin{equation}
\begin{split}
Z_F^i &= (\frac{1}{k}\sum_k Z_L^{i,k}) * Z_V^i \\
P_i &= \sigma(W^cZ_F^i)
\end{split}
\end{equation}
where $W^c \in \mathbb{R}^{d \times d}$ is the weight matrix of the MLP layer and $\sigma$ is the \textit{Softmax} function. For entity $e^i$, $Z_F^i$ indicates the final fused contextual features, $k$ is the token length and $P_i$ is the probability vector.

\textbf{Entity Linking}.
% biffine
The entity linking task desires to extract the relation between any two semantic entities. We use the bi-affine attention~\cite{YueZhang2021EntityRE} for linking prediction, which calculates a score for each relation decision. In particular, for entity $e^i$ and $e^j$, two MLPs are used to project their corresponding features $Z_F^i$ and $Z_F^j$, respectively. After that, a bilinear layer $W^b$ is utilized for the relation score $Score^{i, j}$.
\begin{equation}
\begin{split}
X^i_k &= W^kZ_F^i + b^k \\
X^j_v &= W^vZ_F^i + b^v \\
\text{Score}^{i, j} &= \sigma(X^i_kW^bX^j_v)
\end{split}
\end{equation}
where $W^k, W^v, W^b $ is the parameter weights that $\in \mathbb{R}^{d \times d}$, $b^k$ and $b^v$ is the bias.  $\sigma$ denotes the sigmoid activation function.

% ---------------------------- Experiment ----------------------------
\begin{table*}[t]
  % \begin{threeparttable}
  \begin{center}
  \begin{tabular}{@{\extracolsep{.18cm}}l @{}l @{}l @{}l @{}l @{}l @{}l @{}l @{}l @{}l @{}l @{}l @{}l}
    \toprule
    \textbf{Model} & \textbf{Subject} & \textbf{Test Time} & \textbf{Name} & \textbf{School} & \textbf{\#Exam} & \textbf{\#Seat} & \textbf{Class} & \textbf{\#Student} & \textbf{Grade} & \textbf{Score} & \textbf{Mean} \\
    \midrule
    GraphIE & 94.00 & 100 & 95.84 & 97.06 & 82.19 & 84.44 & 93.07 & 85.33 & 94.44 & 76.19 & 90.26 \\
    TRIE~\cite{PengZhang2020TRIEET} & 98.79 & 100 & 99.46 & 99.64 & 88.64 & 85.92 & 97.94 & 84.32 & 97.02 & 80.39 & 93.21 \\
    VIES & 99.39 & 100 & 99.67 & 99.28 & 91.81 & 88.73 & 99.29 & 89.47 & 98.35 & 86.27 & 95.23 \\
    WatchVIE~\cite{tang2021matchvie} & \textbf{99.78} & 100 & \textbf{99.88} & 98.57 & 94.21 & 93.48 & \textbf{99.54} & 92.44 & 98.35 & 92.45 & 96.87 \\
    StrucTexT~\cite{li2021structext} & 99.25 & 100 & 99.47 & \textbf{99.83} & \textbf{97.98} & 95.43 & 98.29 & 97.33 & \textbf{99.25} & 93.73 & 97.95 \\
    % LayoutLMv3$^\dagger$ & 98.99 & 100 & 99.77 & 99.20 & 100 & 100 & 98.82 & 99.78 & 98.31 & 97.27 & 99.21 \\
    \midrule
    \textbf{Fast-StrucTexT} & 98.39 & \textbf{100} & 99.34 & 99.55 & 96.07 & \textbf{97.22} & 96.73 & \textbf{100} & 95.09 & \textbf{99.41} & \textbf{98.18} \\
    \bottomrule
  \end{tabular}
  \end{center}
  \caption{Entity labeling performance comparison in Chinese on the EPHOIE test set.}
  \label{tab::ephoie}
\end{table*}

\section{Experiment}
In this section, we introduce several datasets used for visual document understanding. We then provide implementation details, including our pre-training and fine-tuning strategies for downstream tasks. We conclude with evaluations of Fast-StrucTexT on four benchmarks, as well as ablation studies.

\subsection{Datasets}
\noindent
\textbf{IIT-CDIP}~\cite{AdamWHarley2015EvaluationOD} is a large resource for various document-related tasks and includes approximately 11 million scanned document pages. Following the methodology of LayoutLMv3, we pre-trained our model on this dataset.

\noindent
\textbf{FUNSD}~\cite{GuillaumeJaume2019FUNSDAD} is a form understanding dataset designed to address the challenges presented by noisy scanned documents. It comprises fully annotated training samples (149) and testing samples (50), and we focused on the semantic entity labeling and linking tasks.

\noindent
\textbf{CORD}~\cite{SeunghyunPark2019CORDAC} is typically utilized for receipt key information extraction, consisting of 800 training receipts, 100 validation receipts, and 100 test receipts. We used the official OCR annotations and an entity-level F1 score to evaluate our model's performance.

\noindent
\textbf{SROIE}~\cite{HuangZheng2019ICDAR2019CO} is a scanned receipt dataset consisting of 626 training images and 347 testing images. Each receipt contains four predefined values: company, date, address, and total. We evaluated our model's performance using the test results and evaluation tools provided by the official evaluation site.

\noindent
\textbf{EPHOIE}~\cite{wang2021towards} is a collection of 1,494 actual Chinese examination papers with a rich text and layout distribution. The dataset is divided into 1,183 training images and 311 testing images, and each character in the document is annotated with a label from ten predefined categories.

\begin{table}[t]
  \centering
  \begin{tabular}{@{\extracolsep{.4cm}}l @{}c}
    \toprule
    Model & FUNSD \\
    \midrule
    BERT~\cite{JacobDevlin2018BERTPO} & 27.65 \\
    SPADE~\cite{WonseokHwang2020SpatialDP} & 41.70 \\
    $\text{StrucTexT}_{BASE}$~\cite{li2021structext} & 44.10 \\
    $\text{LayoutLM}_{BASE}^\dagger$~\cite{xu2020layoutlm} & 48.00 \\
    $\text{LayoutXLM}_{BASE}$~\cite{xu-etal-2022-xfund} & 54.83 \\
    $\text{BROS}_{BASE}$~\cite{hong2022bros} & \textbf{67.63} \\
    \midrule
    \textbf{Fast-StrucTexT} & \underline{67.36} \\
    \bottomrule
  \end{tabular}
  \caption{Entity linking performance comparison on FUNSD dataset. $\text{LayoutLM}_{BASE}^\dagger$ is implemented by BROS. It's worth noticing that our method is $2\times$ faster than BROS.}
  \label{tab::linking}
\end{table}

\subsection{Implementation}
We followed the typical pre-training and fine-tuning strategies to train our model. For all pre-training and downstream tasks, we resized the images along their longer side and padded them to a size of $512 \times 512$. The input sequence is set to a maximum length of 640, with text tokens padded to a length of 512 and image tokens padded to a length of 128.

\begin{table}[t]
  \centering
  \begin{tabular}{@{\extracolsep{.4cm}}l @{}l  @{}l @{}c @{}c}
    \toprule
    Model & FPS$^\ddagger$ & Labeling \\
    \midrule
    Vanilla SA & 70.70 & 88.20 \\
    \midrule
    Linformer~\cite{SinongWang2020LinformerSW} & 73.58 & 83.83 \\
    EAM~\cite{MengHaoGuo2021BeyondSE} & 76.38 & 83.85  \\
    ~ +M\&E  & 91.42 & 86.98  \\
    GFA~\cite{JianWang2022RTFormerED} & 72.82 & 84.08  \\
    ~ +M\&E  & 103.38 & 88.45  \\
    \midrule
    Fast-StrucTexT w/ only SA & 88.88 & 89.59 \\
    Fast-StrucTexT w/o M\&E & 81.85 & 89.94 \\
    \textbf{Fast-StrucTexT} & \textbf{103.50} & \textbf{90.35} \\ 
    \bottomrule
  \end{tabular}
  \caption{Using different transformer design in FUNSD on entity labeling tasks. FPS$^\ddagger$ only calculate the part of the encoder. ``M\&E'' is merging and extension operations for the token.}
  \label{tab::backbone}
\end{table}

\textbf{Model Configurations.} 
The hourglass encoder in our model comprises three M-Blocks and three E-Blocks. Each block consists of a SA and a SCA layer, with 12 heads, in addition to Merging or Extension operations. The encoder architecture is a 12-layer transformer with a hidden size of 768 and an intermediate size of the feed-forward networks, $D_f$, of 3072. The shortening factor $k$ is set to 2 for all stages. The input sequence lengths of the three blocks in M-Blocks are ${256, 128, 64}$, successively, and vice versa in E-Blocks.
We tokenize segment-level text by BERT~\cite{JacobDevlin2018BERTPO}, and trans to text sequence using One-Hot embedding with the vocabulary size $L_v$=30522. To pre-process the image sequence, we use ResNet18~\cite{KaimingHe2015DeepRL} pre-trained on ImageNet~\cite{JiaDeng2009ImageNetAL} to extract ROI features with the size of $C \times H \times W=128 \times 4 \times 64$, followed by a linear project of $D_t$=1024.

\textbf{Pre-training.} 
We pre-train our model on overall IIT-CDIP dataset. The model parameters are randomly initialized. While pre-training, we apply AdamX optimizer with a batch size of 64 for 1 epoch on 8 NVIDIA Tesla A100 80GB GPUs. The learning rate is set as $1 \times 10^{-5}$ during the warm-up for 1000 iterations and the keep as $1 \times 10^{-4}$. We set weight decay as $10^{-4}$ and $(\beta_1=0.9, \beta_2=0.999)$.

\textbf{Fine-tuning on Downstream Tasks.} 
We fine-tune two downstream tasks: entity labeling and entity linking. For entity labeling task on FUNSD, CORD, and SROIE, we set the training epoch as 100 with a batch size of 8 and learning rate of $5 \times 10^{-5}$, $1 \times 10^{-4}$, $1 \times 10^{-4}$, respectively.

\subsection{Comparison with the State-of-the-Arts}
% To the best of our knowledge, there has been no work to design a lightweight model for multi-modal pre-training in structured document understanding to date. 
% Therefore, we
We compare Fast-StrucTexT with BASE scale multi-modal transformer pre-trained models on public benchmarks. We evaluate our model on three benchmark datasets for entity labeling and entity linking tasks with metrics such as Frames Per Second (FPS), Floating Point Operations Per Second (FLOPs), Parameters, and F1 score.

\textbf{Entity Labeling.}
The comparison results are exhibited in Table~\ref{tab::labeling}. Our Fast-StrucTexT achieves state-of-the-art FPS and outperforms the performance of the current state-of-the-art model by 156\%. Fast-StrucTexT achieves a 1.9$\times$ throughput gain and comparable F1 score on the CORD and SROIE entity labeling tasks.
% which sufficiently verifies the superiority of our proposed framework.

To demonstrate the effectiveness of Fast-StrucTexT in Chinese, we pre-train the model in a self-built dataset which consists of 8 million document images in Chinese, and fine-tune the pre-trained model on EPHOIE. Table~\ref{tab::ephoie} illustrates the overall performance of the EPHOIE dataset, where our model obtains the best result with 98.18\% F1-score.

\textbf{Entity linking}.
As shown in Table~\ref{tab::linking}, we compare Fast-StrucTexT with several state-of-the-art methods on FUNSD for entity linking. Compared with BROS~\cite{hong2022bros}, our method achieves a comparable performance with 2$\times$ speed in Table~\ref{tab::labeling}.

\subsection{Ablation Studies}
We conduct ablation studies on each component of the model on the FUNSD and SROIE datasets, including the backbone, pre-training tasks, pooling strategy, and shorten factor. At last, we evaluate the cost of our proposed hourglass transform with various sequence lengths.

\textbf{Backbone.} 
To prove the pre-trained Fast-StrucTexT can obtain state-of-the-art efficiency and performance. We replaced a variety of popular lightweight backbones~\cite{MengHaoGuo2021BeyondSE,SinongWang2020LinformerSW,JianWang2022RTFormerED} for evaluation. As shown in Table~\ref{tab::backbone}, Fast-StrucTexT can achieve the highest performance and FPS. The point here is that transformer takes responsibility for token feature representation, which could benefit from large model size. It is the reason why those lightweight architectures lead to worse performance. Particularly, we consider that those methods only reduce time complexity, but it is not considered that multi-modal interaction for visual document understanding. Therefore, we integrate multi-modal interaction into the design of our model.

In Table~\ref{tab::backbone}, the model shows better efficiency and performance than other lightweight encoders and achieves FPS = 103.50, F1 = 90.35\%. Comparing ``Ours w/o M\&E'' and ``Ours'', we can obverse that merging and extension operations are not only efficient but also can improve performance. Specifically, it obtains $1.5\times$ the throughput of ''w/o M\&E'' settings, and 0.41\% improvement in the labeling task.

\textbf{Pooling.} 
Table~\ref{tab::pooling} gives the results of different merging strategies for entity labeling task on FUNSD. GlobalPool is a form to directly merge all token-level features into segment-level before encoding. AvgPool is our merging strategy without cross-modal guidance. DeformableAttention~\cite{xia2022vision} attempts to learn several reference points to sample the sequence. The experimental results show the effectiveness of our token merging method.

\begin{table}[t]
  \centering
  \begin{tabular}{@{\extracolsep{.4cm}}l @{}l  @{}l @{}c @{}c}
    \toprule
    Method & FPS$^\ddagger$ & FUNSD \\
    \midrule
    GlobalPool & \textbf{120.33} & 86.62 \\
    AvgPool & 107.46 & 88.01 \\
    DeformableAttention & 83.14 & 88.57 \\
    \midrule
    \textbf{Fast-StrucTexT}& 103.50 & \textbf{90.35} \\
    \bottomrule
  \end{tabular}
  \caption{Ablation of token merging strategies.}
  \label{tab::pooling}
\end{table}

\begin{figure}
    \centering
    \includegraphics[width=1.0\linewidth]{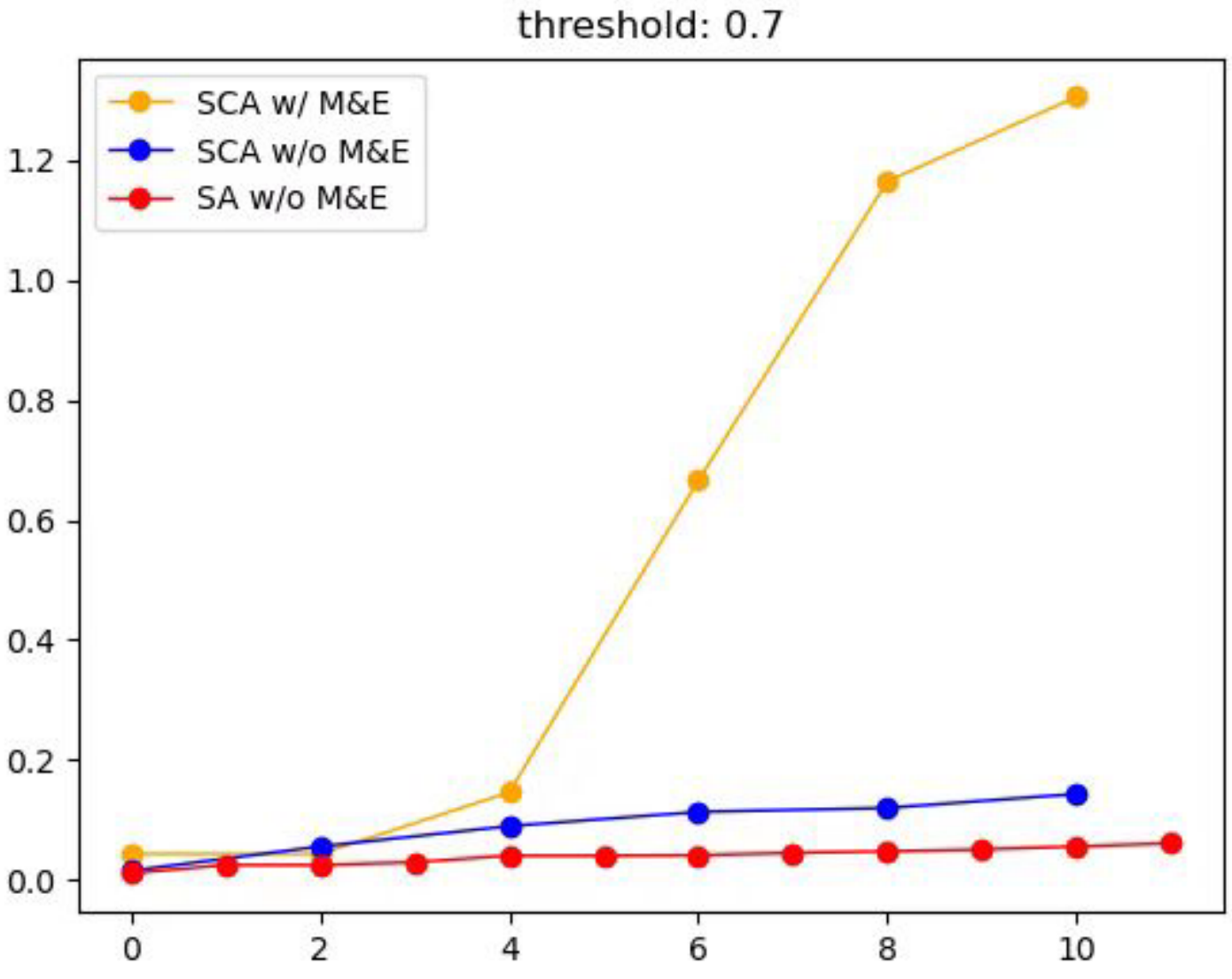}
    \caption{The cumulative proportions of the filted tokens which probability beyond 0.7 in attention maps of transformers. SCA has better modal interaction ability.}
    \label{fig:count}
\end{figure}

\textbf{Shorten factor.} 
We extend the ablation study on the hyper-parameter $k$ in Table~\ref{tab:k}. We have evaluated multiple benchmarks and have established that the setting $k$=2  is the best trade-off for document understanding.  In addition, we can adjust the $k$ in the fine-tuning stage. Nevertheless, ensuring consistency of $k$ during pre-training and fine-tuning can take full advantage of knowledge from pre-trained data. Our framework supports multi-scale pre-training with a list of $k$ factors to handle different shortening tasks.

\begin{table}[!ht]
    \setlength{\tabcolsep}{0.9mm}{
    \renewcommand{\arraystretch}{1}
        \centering
        \begin{tabular}{l|ccccc}
        \toprule
        $k$ & SROIE-F1 & FUNSD-F1 & CORD-F1 & FPS & FLOPs \\
        \midrule
        1 & 97.46 & \textbf{90.54} &  97.15 & 81.85  & 46.48G  \\
        \textbf{2} & \textbf{97.55} & 90.35 & \textbf{97.15} & 103.50 & 19.85G   \\
        4 & 95.88 & 88.69 &  96.85 & 117.69 & 12.46G   \\
        8 & 92.22 & 83.07 &  93.85 & 133.48 & 10.14G  \\
    \bottomrule
    \end{tabular}
    \caption{Ablation study of the various $k$.}
    \label{tab:k}}
\end{table}

\textbf{Sequence length.} 
Our token merging can adapt to the arbitrary length of the input. The ratio of merging is adjustable and determined by the value of $k$ and the number of M-blocks and E-blocks. Referring to Table~\ref{tab:speed}, we investigate the ability of our method with various sequence lengths. The experimental results show that the speed and computation gains become more pronounced with the sequence length increasing.

\begin{table}[!ht]
\setlength{\tabcolsep}{0.9mm}{
\renewcommand{\arraystretch}{1}
    \centering
    \begin{tabular}{l|cc|cc}
    \toprule
     Seq. &  V.T.   & Ours     &  V.T.      &  Ours   \\
     Len. &  (FPS) & (FPS)    & (FLOPs)   &  (FLOPs)\\
    \midrule
       512  & 58.39  & 77.20 (+32\%)  & 48.36G & 19.91G (-58\%) \\
       1024 & 30.25  & 52.86 (+74\%) & 106.39G & 41.57G (-60\%) \\
       2048 & 13.46  & 24.76 (+84\%) & 251.44G & 90.12G (-64\%) \\
       4096 & 4.68   & 10.59 (+126\%) & 657.50G & 208.16G (-68\%) \\
       8192 & 1.22   & 3.82 (+213\%) & 1933.49G & 527.95G (-72\%) \\
    \bottomrule
    \end{tabular}
    \caption{Ablation study of sequence length, where V.T. represents Vanilla Transformer.}
    \label{tab:speed}}
\end{table}

As shown in Fig~\ref{fig:count}, we conducted an ablation study on the hourglass architecture and SCA module. We calculated the number of more than 0.7 in each layer of the attention map and then accumulated it layer by layer to observe the utilization of each token. Compared with the yellow line and the blue line, the number of highly responsive areas in the later layer is significantly increased, which can show that after aggregating the multi-granularity semantics, each token has sufficient ability to express more information. The blue line is slightly higher than the red line, which indicates that our SCA has better modal interaction ability and lower computational capacity than SA.

% ---------------------------- Conclusion ----------------------------
\section{Conclusion}
In this paper, we present Fast-StrucTexT, an efficient transformer for document understanding task. Fast-StrucTexT significantly reduces the computing cost through hourglass transformer architecture, and utilizes multi-granularity information through modality-guided dynamic token merging operation. Besides, we propose the Symmetry Cross-Attention module to enhance the multi-modal interaction and reduce the computational complexity. Our model shows state-of-the-art performance and efficiency on four benchmark datasets, CORD, FUNSD, SROIE, and EPHOIE.
% ---------------------------- Done ----------------------------

\bibliographystyle{named}
\bibliography{ijcai23}

\end{document}